\documentclass[runningheads]{llncs}
\usepackage[T1]{fontenc}

\usepackage{graphicx,verbatim}
\usepackage{amsmath,amssymb}
\usepackage{subcaption}
\usepackage[dvipdfmx]{xcolor}
\DeclareMathOperator{\artanh}{artanh}

\begin{document}
\title{Multi-Scale Representation \\ of Follicular Lymphoma Pathology Images \\ in a Single Hyperbolic Space}

\begin{comment}
\end{comment}

\author{Kei Taguchi\inst{1} \and Kazumasa Ohara\inst{1} \and Tatsuya Yokota\inst{1} \and Hiroaki Miyoshi\inst{2} \and \\ Noriaki Hashimoto\inst{3} \and Ichiro Takeuchi\inst{4} \and Hidekata Hontani\inst{1}}
\authorrunning{K. Taguchi et al.}
\institute{Nagoya Institute of Technology \\ 
\email{k.taguchi.015@nitech.jp, k.ohara.674@nitech.jp, t.yokota@nitech.ac.jp, hontani@nitech.ac.jp} \\
\and
Kurume University \\
\email{miyoshi\_hiroaki@med.kurume-u.ac.jp} \\
\and
RIKEN \\
\email{noriaki.hashimoto.jv@riken.jp} \\
\and
Nagoya University \\
\email{itakeuchi73@gmail.com} \\
}

\maketitle

\begin{abstract}
We propose a method for representing malignant lymphoma pathology images, from high-resolution cell nuclei to low-resolution tissue images, within a single hyperbolic space using self-supervised learning. 
To capture morphological changes that occur across scales during disease progression, our approach embeds tissue and corresponding nucleus images close to each other based on inclusion relationships. 
Using the Poincar\'e ball as the feature space enables effective encoding of this hierarchical structure. 
The learned representations capture both disease state and cell type variations. 
\end{abstract}

\section{Introduction}
\subsection{Objective}
In this study, we propose a method to represent the multi-scale hierarchical structure from individual cell nucleus images to large-scale tissue images observed in pathology images of malignant lymphoma \cite{flcontent} using self-supervised learning within a single hyperbolic space.
The morphology of biological tissues often changes in coordination with disease progression, where the morphology of cell nuclei and the morphology of tissues change in a coordinated manner \cite{koga2024study,who2008}.
To reflect such cross-scale morphological changes in image representation, we perform representation learning based on the inclusion relationships between images (see Fig. \ref{fig:pretrain_model} (Left)).

Most existing multi-scale representations of pathological images encode cell-level and tissue-level images using different encoders. 
The hierarchical scale structure is typically represented by inputting the encoded results of cell-level images into the tissue-level encoder, as in models such as U-Net\cite{ronneberger2015u} and HIPT\cite{chen2022scaling}. 
Cancer subtype classification is usually performed using tissue-level images (e.g., via multiple instance learning\cite{ilse2018attention}), while it is generally difficult to classify cancer subtypes solely from cell-level images. 
Although both tissue structures and cell nuclei change during cancer progression, not all tissue regions in a whole slide image (WSI) exhibit subtype-specific alterations, nor do all nuclei within a cancerous region show subtype-specific features. 
In supervised learning of hierarchical representations, subtype classification becomes achievable at the tissue level, which in turn limits the extent to which cell-level representations reflect subtype-related features. 
Even with self-supervised learning approaches such as HIPT, cell-level representation is trained independently from other scale images, making it difficult for representations of cell-level images to incorporate cancer subtype variations.

To address this, we propose a self-supervised learning method that represents cross-scale morphological changes in pathological images by embedding multi-scale images—from cell nucleus images $ ( 56 \times 56 ) $ to tissue images $ ( 4096 \times 4096 ) $—into a single hyperbolic space using a shared encoder.
This embedding reflects not only the cross-scale appearance similarities, but also the inclusion-based hierarchy between image patches of different scales. 
That is, when one image contains another, the encoder learns to embed these two images to proximal positions in the feature space.
The ability to represent such nested structures enables cell nucleus image representations to reflect subtype differences through their interaction with tissue image representations.

When embedding the inclusion-based hierarchy between image patches of different scales, using an Euclidean space as the feature space results in a lack of capacity, leading to an inability to assign appropriate embeddings \cite{linial1995geometry}. 
The method proposed in \cite{ge2023hyperbolic} uses a hyperbolic space for the feature space for the embedding of the hierarchical structure. 
It is known that one can embed tree structures with minimal distortion in a hyperbolic space \cite{gromov1987hyperbolic,li2023euclideanspaceevilhyperbolic,sarkar2011low}.

During the learning process, the representation of each global image is directly affected by the representations of local images included in it, while local image representations are affected by the representations of global images.
In this study, the smallest-scale local image is an image containing only a single cell nucleus.
If some cell nuclei are specific to some disease, that nucleus images will be embedded to near the tissues image of that disease.

\subsection{Preliminaries}
\subsubsection{Hyperbolic Space}
The proposed method adopts Poincar\'e ball \cite{ganea2018hyperbolic,gromov1987hyperbolic,nickel2017poincare,tifrea2018poincar} as the feature space.
Unlike Euclidean space, the volume of Poincar\'e ball grows exponentially with the distance from the origin. 
The Poincar\'e ball allows embedding a greater number of points as they move farther from the origin, so it is well-suited for representing hierarchical data.

The Riemannian metric of the $d$-dimensional Poincar\'e ball $\mathbb{B}^d_\kappa:=\{\boldsymbol{z} \in \mathbb{R}^d | \kappa\|\boldsymbol{z}\|^2 < 1\}$, where $\kappa > 0$ is the (negative) curvature, is given as follows:

\begin{equation}\label{poincare_metric}
  g_{\mathbb{B}}(\boldsymbol{z}) = \frac{4}{\left(1 - \kappa\|\boldsymbol{z}\|^2\right)^2}g_{\mathbb{E}}
\end{equation}
where $\boldsymbol{z} \in \mathbb{B}^d_\kappa$ is the position vector with the origin as the center of the Poincar\'e ball, and $g_{\mathbb{E}}$ is the standard metric of Euclidean space \cite{durrant2023hmsn,ge2023hyperbolic}.

In the Poincaré ball model, the radius $r$ and the curvature $\kappa$ satisfy the following relation:
\begin{equation}\label{rad_cur}
  \kappa = \frac{1}{r^2}
\end{equation}
In this study, the curvature and radius of the Poincar\'e ball are fixed at $\kappa=1$, $r=1$.

As $\|\boldsymbol{z}\|$ approaches 1, the coefficient of $g_{\mathbb{E}}$ increases, causing movements to become larger as $\boldsymbol{z}$ moves away from the center.
The geodesic distance on the Poincar\'e ball between two points $\boldsymbol{z}_1, \boldsymbol{z}_2 \in \mathbb{B}^d_\kappa$ can be computed as follows:
\begin{equation}\label{poincare_dist}
  d_{\mathbb{B}}(\boldsymbol{z}_1, \boldsymbol{z}_2) = \frac{2}{\sqrt{\kappa}} \artanh \left( \sqrt{\kappa} \| -\boldsymbol{z}_1 \oplus_\kappa \boldsymbol{z}_2 \|  \right)
\end{equation}

The exponential map of the Poincar\'e ball, which projects Euclidean embeddings onto the Poincar\'e ball and is centered at the origin, is given by:
\begin{equation}\label{exp_map}
  \exp^{\kappa}_{\boldsymbol{z}}(\boldsymbol{v}):= \boldsymbol{z} \oplus_\kappa \left(\tanh\left( \frac{\sqrt{\kappa} \|\boldsymbol{v}\|}{1-\kappa\|\boldsymbol{z}\|^2}\right) \frac{\boldsymbol{v}}{\sqrt{\kappa}\|\boldsymbol{v}\|}\right)
\end{equation}
where $\boldsymbol{z} \in \mathbb{B}^d_\kappa$ and $\boldsymbol{v}\in T_z\mathbb{B}^d_\kappa \approx \mathbb{R}^d$.
This exponential map converts a Euclidean latent space to a hyperbolic one. 
$\oplus_\kappa$ denotes M\"obius addition, a differentiable operation defined as follows:
\begin{equation}\label{m_add}
  x \oplus_\kappa y = \frac{(1 + 2\kappa \langle x, y \rangle + \kappa \|y\|^2)x+ (1 - \kappa \|x\|^2)y}{1 + 2\kappa \langle x, y \rangle + \kappa^2 \|x\|^2 \|y\|^2}
\end{equation}

In the proposed method, an embedding $\boldsymbol{z} \in \mathbb{B}^d_\kappa$ on the Poincaré ball is assigned to each prototype $\boldsymbol{c}$ placed on the boundary $\partial \mathbb{B}^d_\kappa$ of the Poincaré ball. 
The boundary $\partial \mathbb{B}^d_\kappa$ corresponds to the sphere $\mathbb{S}^{d-1}$ and is not included in the Poincaré ball. To evaluate the assignment of $\boldsymbol{z}$ to a prototype $\boldsymbol{c}$, we employ the Busemann function \cite{bridson2013metric,busemann2012geometry,ghadimi2021hyperbolic}:
\begin{equation} \label{busemann}
  b_{\boldsymbol{c}}(\boldsymbol{z}) = \log \frac{\|\boldsymbol{c} -\boldsymbol{z}\|^2}{1-\|\boldsymbol{z}\|^2}
\end{equation}
\section{Method}
\subsection{Proposed Method}
\begin{figure}[tb]
  \centering
  \includegraphics[width=0.85\linewidth]{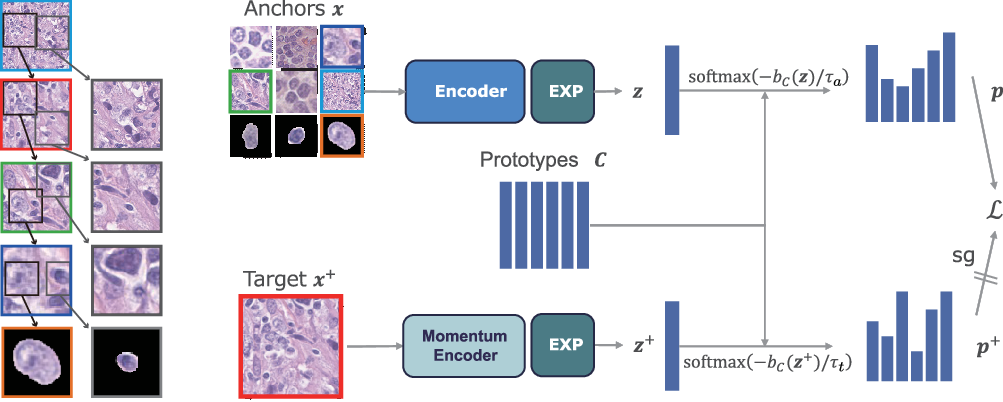}
  \caption{
    (Left) Inclusion relationships between images, from a tissue image ($4096\times 4096$) down to a cell nucleus image ($56\times 56$). (Right)Self-supervised learning using hyperbolic embeddings and prototypes. Image pairs with inclusion relationships, as shown on the left, are selected as anchor and target.
  }
  \label{fig:pretrain_model}
\end{figure}

The proposed method adopts the loss function used in Hyperbolic Masked Siamese Networks (HMSN) \cite{durrant2023hmsn}.
HMSN uses prototypes \cite{assran2022masked,caron2020unsupervised} to structure representations in hyperbolic space. 
In the proposed method, the prototypes placed on the boundary of the Poincar\' {e} ball \cite{durrant2023hmsn}, guide the embeddings of images based on the inclusion relationships between the global and local morphology. 

The proposed method extracts tissue images, $\boldsymbol{x}_i$, from WSIs. 
For each $\boldsymbol{x}_i$, one target view and $M$ anchor views with inclusion relationships are generated. 
The target view is denoted as $\boldsymbol{x}_i^+$ and anchor views as $\boldsymbol{x}_{i,m}$ where $m=1, 2, \cdots, M$.
In contrast to HMSN \cite{durrant2023hmsn}, where masking and conventional image augmentation are used to generate the anchor and targets from an image, our method constructs anchor-target pairs from images of different spatial scales based on their inclusion relationships (see Fig. \ref{fig:pretrain_model} (Right)). 
This enables cross-scale learning of hierarchical features. 
Here, the scale of an image refers to the crop size when extracting it from $\boldsymbol{x}_i$. 
Each is mapped to the Poincar\'e ball using Equation (\ref{exp_map}) to obtain $\boldsymbol{z}_i^+ \in \mathbb{B}^d_\kappa$ and $\boldsymbol{z}_{i,m} \in \mathbb{B}^d_\kappa$ respectively.
$K$ learnable prototypes $\boldsymbol{c}_k \in \mathbb{R}^d$ are placed on the Poincar\'e ball boundary $\partial \mathbb{B}^d_\kappa$, and the assignment $b_{\boldsymbol{c}_k}(\boldsymbol{z})$ from $\boldsymbol{z}$ to each $\boldsymbol{c}_k$ are computed using the Busemann function shown in Equation (\ref{busemann}).

Let the set of $K$ computed assignments be denoted by a $K$-dimensional vector, $\boldsymbol{b}_{\boldsymbol{c}}(\boldsymbol{z}) = \left[b_{\boldsymbol{c}_1}(\boldsymbol{z}), \ldots, b_{\boldsymbol{c}_K}(\boldsymbol{z})\right]^{\top}$. Each feature vector $\boldsymbol{z}$ is represented by using the prototype assignment vector $\boldsymbol{p} \in \Delta_K$ computed as follows:
\begin{equation}\label{hmsni_p}
\boldsymbol{p}:=\mathrm{softmax}\left(\frac{-\boldsymbol{b}_{\boldsymbol{c}}(\boldsymbol{z})}{\tau_{*}}\right)
\end{equation}
Following HMSN \cite{durrant2023hmsn}, we use different temperature parameters for the target and anchor views in the soft assignment step.
Let $\tau_{t}$ denote the temperature for the target view and $\tau_{a}$ denote the temperature for the anchor view.
We set $\tau_{t} = 0.0625$ and $\tau_{a} = 0.25$ in all experiments.
Among the images used for encoding, cell nucleus images have the smallest scale. 
As for cell nucleus images, they are generated by segmenting cell nuclei, cropping each nucleus centered at its centroid, and setting the pixel values outside the cell nucleus to zero aiming to isolate the morphological features of cell-images.

The distance between $\boldsymbol{p}_i^+$ obtained from the target view and $\boldsymbol{p}_{i,m}$ obtained from the anchor view is evaluated using the cross-entropy $\frac{1}{MB}\sum_{i=1}^{B}\sum_{m=1}^{M}H(\boldsymbol{p}_i^+, \boldsymbol{p}_{i,m})$.
To ensure that all prototypes are utilized evenly, a regularization term is introduced to maximize the entropy $H(\bar{\boldsymbol{p}})$ of the average the prototype assignment vector across all anchor views, where $\bar{\boldsymbol{p}} = \frac{1}{MB}\sum_{i=1}^{B}\sum_{m=1}^{M}\boldsymbol{p}_{i, m}$ \cite{assran2021semi}.
Additionally, to encourage anchor views to be assigned to specific prototypes, the term $\frac{1}{MB}\sum_{i=1}^{B}\sum_{m=1}^{M}H(\boldsymbol{p}_{i,m})$ is minimized.
The encoder and prototypes are trained to minimize the following objective function:
\begin{equation}\label{loss}
\mathcal{L} = \frac{1}{MB}\sum_{i=1}^{B}\sum_{m=1}^{M}H(\boldsymbol{p}_i^{+}, \boldsymbol{p}_{i,m}) - \lambda H(\bar{\boldsymbol{p}}) + \beta \frac{1}{MB}\sum_{i=1}^{B}\sum_{m=1}^{M} H(\boldsymbol{p}_{i, m})
\end{equation}
where $\lambda$ and $\beta$ are hyperparameters.
\subsection{Dataset and Implementation}
\subsubsection{Dataset}
From a malignant lymphoma database, approximately 5,000 WSIs with subtype diagnoses were selected, and from these, around 150,000 tissue images were obtained by randomly extracting $4096 \times 4096$ tissue regions from histological sections.
The selection of WSIs was not restricted by subtype, and the extraction of tissue images did not prioritize specific histological features, such as follicles, which could contribute to disease classification.
It should be noted that the number of subtypes included in the training data was more than $70$, and many of the sampled images originated from regions within WSIs that are less likely to contribute to subtype classification. 

The dataset included a significant number of tissue images from subtypes other than Reactive, Follicular Lymphoma (FL), and Diffuse Large B-Cell Lymphoma (DLBCL), which were later used for validation.
The following experiments are conducted using images of FL, DLBCL, and Reactive cases. 

To generate target and anchor views, tissue images were cropped from random positions within the $4096 \times 4096$ tissue images. Three different crop sizes were used: $224 \times 224$, $112 \times 112$, and $56 \times 56$.
During training, disease subtype labels were not used and were referenced only for evaluation.
To obtain individual cell nucleus images from the tissue images, HoVer-Net \cite{graham2019hover} was applied for cell nucleus segmentation.
Regardless of nucleus size, each nucleus image was extracted as a fixed-size $56 \times 56$ patch, with the surrounding region outside the nucleus set to zero.
In total, approximately 2.9 billion cell nucleus images were generated and used.

Cell nuclei within follicles play a crucial role in diagnosing Reactive and FL cases \cite{who2008}. 
Therefore, cell type labels were assigned to the nuclei inside follicles in WSIs of Reactive and FL cases by an expert, categorizing them into seven classes: Round-Medium, Centrocyte, Centroblast, Immunoblast, Cleaved-Large, FDC (Follicular Dendritic Cell), and Others. 
Since DLBCL does not contain follicles, nuclei from randomly selected tissue images were labeled by the expert pathologists.
These annotations were not used during self-supervised learning but were utilized for analyzing embeddings in the Poincar\'{e} ball, as well as for training and evaluating a cell type classifier for a downstream task.
The number of labeled cell nucleus images for each cell type is as follows. Round-Medium has 561 images, Centrocyte has 683 images, Centroblast has 239 images, Immunoblast has 29 images, Cleaved-Large has 123 images, and FDC has 93 images.
When evaluating a cell type classifier for a downstream task, we created another dataset consistings of four nucleus types with relatively abundant data: Round-Medium (307 samples), Centrocyte (297), Centroblast (49), and Cleaved-Large (67). 
A total of 720 annotated samples were used for a 4-fold cross-validation without altering the class distribution.
\subsubsection{Implementation}
In this study, the encoder is based on ResNet34 \cite{he2016deep}, with the final linear layer replaced by a multilayer perceptron, and an additional exponential mapping shown in Equation (\ref{exp_map}) applied.
The dimension of the Poincar\'e ball is set to $d=64$.
The loss function coefficients in Equation (\ref{loss}) are set to $\lambda=10$ and $\beta=0.25$.
The target temperature in Equation (\ref{hmsni_p}) is set to $\tau_{t}=0.0625$, and the anchor temperature is set to $\tau_{a}=0.25$.
The number of prototypes is set to $K=512$. 
The model was trained for $500$ epochs with a batch size of $1024$.
We used RiemannianAdam \cite{becigneulriemannian} for optimizer. 
In training, the first 50 epochs are used as warm-up epochs, during which the learning rate is linearly increased to 1e-4. After that, a cosine scheduler is used to decrease the learning rate to 1e-5.
For cell nucleus type classification, we employ 4-fold cross-validation on 720 annotated cell nucleus images, preserving class ratio. 
In each fold, we use 540 images for training and 180 images for validation.

\subsubsection{Limitations}
In our current setup, nuclear images are created by extracting a $56\times 56$ patch centered on each segmented nucleus, with all pixels outside the nuclear region set to zero.
This hard-masked background, while helpful for isolating nuclear morphology, introduces an artificial boundary that could bias representation learning. 
Moreover, the nucleus segmentation is performed by HoVer-Net\cite{graham2019hover}, and inaccuracies in segmentation—particularly in dense or overlapping regions—could introduce noise or bias into the downstream representation space. 
Future work should consider more nuanced strategies, such as soft masking, confidence-aware sampling, or incorporating cytoplasmic features where feasible. 

\section{Experiments and Results}
\subsection{Distribution on the Poincar\'e Ball}
\begin{figure}[tb]
  \centering
  \includegraphics[width=0.85\linewidth]{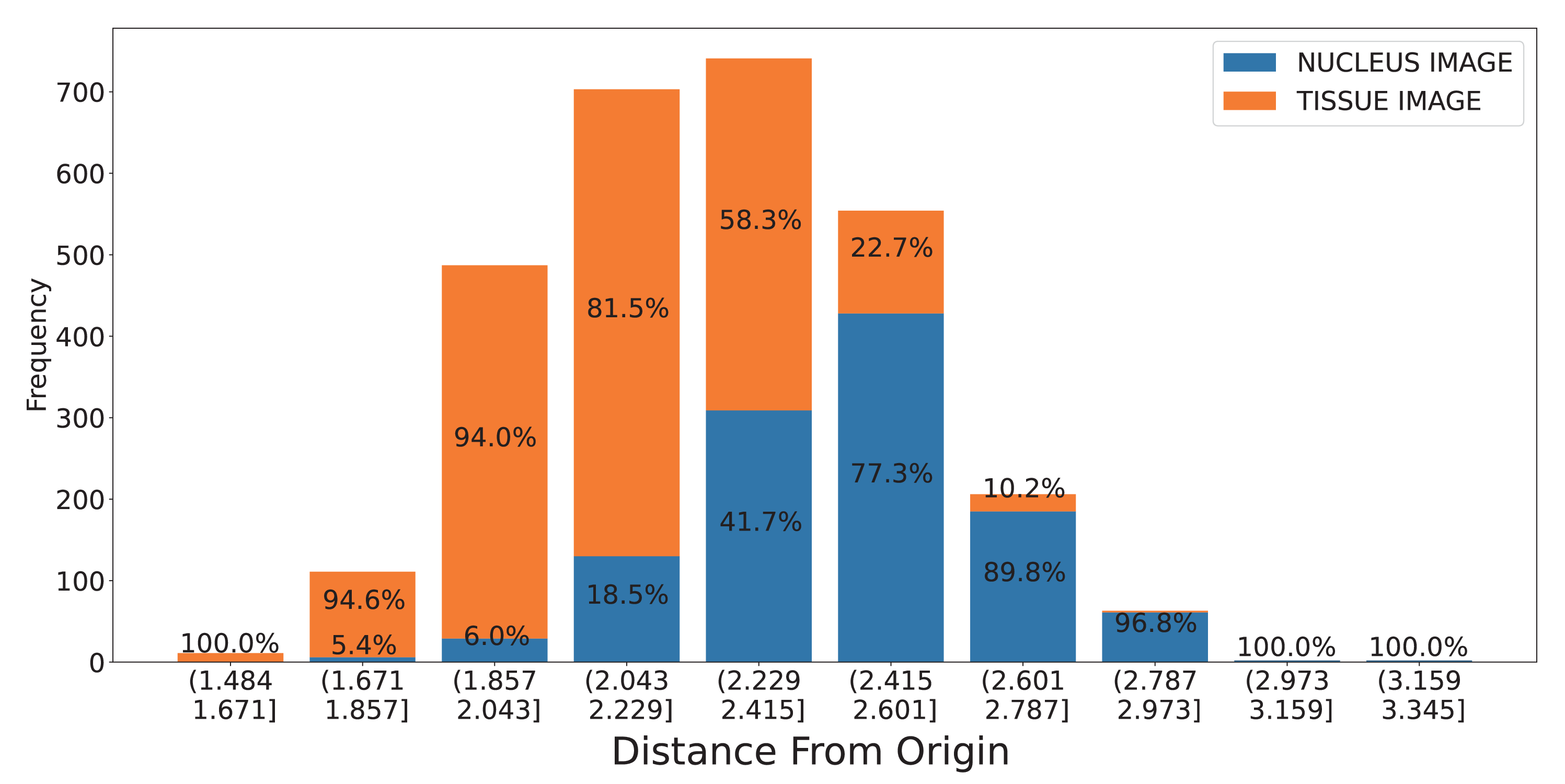}
  \caption{Distance of tissue and cell nucleus image representations from the origin: 
  Nucleus images, represented with hard-masked backgrounds, tend to be located near the boundary. Tissue images reside closer to the origin. }
  \label{fig:freq}
\end{figure}
\begin{figure}[tb]
  \centering
  \includegraphics[width=80mm]{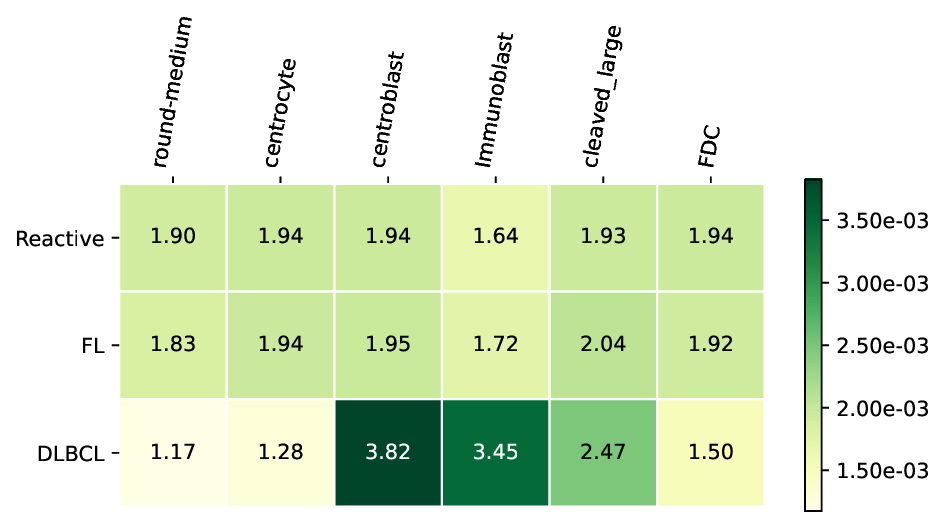}
  \caption{
    Heatmap showing the assignment degrees between subtypes (columns) and annotated nucleus types (rows). Higher values indicate that the corresponding tissue and nucleus images tend to be assigned to similar prototypes. These results suggest subtype-specific patterns in prototype usage, particularly for centroblasts and immunoblasts in DLBCL. 
  }
  \label{fig:sub_cell}
\end{figure}
We analyze the distribution of tissue and nuclear image representations on the Poincaré ball after learning. 
Representations are expected to form a continuous hierarchy by distance from the origin \cite{nickel2018learningcontinuoushierarchieslorentz,yang2023hyperbolicrepresentationlearningrevisiting}. 
Fig. \ref{fig:freq} shows the discretized distribution of nuclear and other tissue image representations by their distance from the origin.
In the figure, the horizontal axis represents the geodesic distance from the origin, while the vertical axis indicates the frequency at each geodesic distance. 
The numerical values shown are percentages. The proportion of cell nucleus images increases as the distance from the origin increases. 
The authors believe that this result reflects the multi-scale hierarchical structure formed by the inclusion relationships.

\subsection{Distribution of Tissue and Cell Nucleus Images}
To investigate how the learned prototype assignments reflect subtype-specific patterns, we calculate the prototype assignment vector $\boldsymbol{p}$ between each subtype and each cell nucleus type. 
Since the embedded images are labeled with disease subtypes, we compute the mean assignment vector for each subtype. 
Let the average prototype assignment vector for Reactive, FL, and DLBCL be denoted by $\boldsymbol{p}_{\text{Reactive}}, \boldsymbol{p}_{\text{FL}}, \boldsymbol{p}_{\text{DLBCL}} \in \Delta_{K}$.
If tissue images of the same disease subtype are assigned to specific prototypes, the corresponding components of these average prototype assignment vector, $\boldsymbol{\bar{p}}_{\mathrm{subtype}}$ will have higher values.
Some cell nucleus images are also labeled with their nucleus type. 
The mean vector for each nucleus type, $\boldsymbol{\bar{p}}_{\mathrm{cell}}$ is calculated in a same manner.
The degree to which tissue images of a particular disease subtype and images of a specific cell nucleus type are assigned to the same prototype is evaluated by the inner product of their average prototype assignment vector
The assignment degree is defined as the inner product $\boldsymbol{\bar{p}}_{\text{subtype}} \cdot \boldsymbol{\bar{p}}_{\text{cell}}$, which quantifies the similarity between the two distributions. 
A higher value indicates that tissue and nucleus images of the given types tend to be assigned to the same prototypes. 

The resultant degree of assignment is shown in Fig. \ref{fig:sub_cell}. 
It should be noted that, during the self-supervised learning, a large number of cases from subtypes other than the three shown here (Reactive, FL, and DLBCL) were also used, and subtype labels were not referenced during training. 
The progression of cancer follows the order of Reactive → FL → DLBCL.

According to the Fig. \ref{fig:sub_cell}, among the different cell types, Centroblasts, Immunoblasts, and Cleaved Large cells increase with cancer progression, while Centrocytes decrease. 
This change is particularly pronounced in DLBCL. Notably, the increase in Centroblasts and the decrease in Centrocytes with cancer progression aligns with the WHO grading criteria for FL \cite{who2008}.

\subsection{Cell nucleus type classification}
The classifier architecture is based on the pretrained model, specifically ResNet34 \cite{he2016deep} with an added multilayer perceptron.
The representation before the exponential mapping is $L_2$ normalized, followed by an additional linear layer.
We compare this architecture with a randomly initialized model without pretraining and a pretrained model using the proposed representation learning.

Table \ref{tab:cell_class} presents the classification performance.
Cell nucleus type annotation is challenging even for expert pathologists \cite{koga2024study}.
Achieving high classification performance of the cell type classification is stil difficult.
Though, pretraining improved performance, confirming the robustness of the learned representation to label noise.
\begin{table}[tb]
  \centering
  \caption{Results of cell nucleus type classification.}
    \label{tab:cell_class}
    \begin{tabular}{|c|c|c|c|c|}
        \hline
        & Accuracy & F1 score (macro) & Precision (macro) & Recall (macro) \\ \hline
        w/o pretrained & 0.682 & 0.631 & 0.657 & 0.617 \\ \hline
        w/ pretrained & \bf 0.735 & \bf 0.714 & \bf 0.736 & \bf 0.702 \\ \hline
    \end{tabular}
\end{table}
\section{Conclusion}
We proposed a method to represent malignant lymphoma pathology images in a single Poincaré ball using self-supervised learning, covering images from cell nuclei to tissue scale. 
By leveraging inclusion relationships between images, our method embeds tissue and contained nucleus images in close proximity. 
Future work includes extending the analysis to a broader range of lymphoma subtypes and statistical significance tests.

\bibliographystyle{splncs04}
\bibliography{mybibliography}
\end{document}